# Adaptive knowledge transferring with switching dual-student framework for semi-supervised medical image segmentation

Hoang-Thien Nguyen [a,1], Thanh-Huy Nguyen [b,1], Ba-Thinh Lam [a], Vi Vu [c], Bach X. Nguyen [c], Jianhua Xing [d], Tianyang Wang [e], Xingjian Li [b], Min Xu [b,*]

[a] PASSIO Lab, North Carolina A&T State University, Greensboro, 27411, NC, USA
[b] Carnegie Mellon University, Pittsburgh, 15213, PA, USA
[c] Ho Chi Minh University of Technology, Ho Chi Minh, 70000, Viet Nam
[d] University of Pittsburgh, Pittsburgh, 15260, PA, USA
[e] University of Alabama at Birmingham, Birmingham, 35294, AL, USA

ABSTRACT

Teacher-student frameworks have emerged as a leading approach in semi-supervised medical image segmentation, demonstrating strong performance across various tasks. However, the learning effects are still limited by the strong correlation and unreliable knowledge transfer process between teacher and student networks. To overcome this limitation, we introduce a novel switching Dual-Student architecture that strategically selects the most reliable student at each iteration to enhance dual-student collaboration and prevent error reinforcement. We also introduce a strategy of Loss-Aware Exponential Moving Average to dynamically ensure that the teacher absorbs meaningful information from students, improving the quality of pseudo-labels. Our plug-and-play framework is extensively evaluated on 3D medical image segmentation datasets, where it outperforms state-of-the-art semi-supervised methods, demonstrating its effectiveness in improving segmentation accuracy under limited supervision.

## 1. Introduction

Analyzing segmentation from computer tomography (CT) or magnetic resonance imaging (MRI) is essential for numerous clinical applications. Current methods for medical image segmentation focus on exploiting annotated images, which require a large number of human-labeled datasets. However, in practice, pixel-wise annotation of CT or MRI images by specialists is a labor-intensive and time-consuming process. To alleviate the burden of labeling, many studies focused on Semi-Supervised Medical Image Segmentation (SSMIS), leveraging a large number of unlabeled images alongside limited annotated images during training.

To learn from the unlabeled images, most SSMIS approaches resort to generating pseudo labels or consistency-based supervision via a surrogate annotator. This inspires the design of Mean-Teacher [1], which remains a dominant architecture in recent state-of-the-art methods. Mean-Teacher (MT) involves a teacher model generating pseudo supervision signals for a student model, while the teacher itself is iteratively updated using an Exponential Moving Average (EMA) of the student's weights. Despite significant advancements in MT-based SSMIS research, enhancing the quality of pseudo-labels or supervisory signals remains an active and ongoing research challenge. Some approaches consider post-hoc pseudo-label refinement based on prediction confidence [2–5] or stability [6,7]. These approaches primarily intervene at the output stage rather than enhancing the quality of the teacher model itself. A key limitation is that, a low-quality teacher model is not able to correct inaccurate pseudo labels, but instead has to discard them, leading to inefficient learning from uncertain samples. Dual-Student [8] assumed that the teacher model is a combination of a set of learned students, which implies that the teacher's performance is bounded below by that of the students and proposed dual-student architecture to address this problem. Some other works focus on multi-view image augmentation [9–14] or multi-task [15,16] learning to facilitate effective consistency learning. These approaches usually require domain-specific knowledge or strategies to achieve optimal performance.

* Corresponding author.
*E-mail addresses:* thienhaiquan123@gmail.com (H.-T. Nguyen), bihuyz@gmail.com (T.-H. Nguyen), thinhlb99@gmail.com (B.-T. Lam), vulanvi22@gmai.com (V. Vu), bach.nguyenspring@hcmut.edu.vn (B.X. Nguyen), xing1@pitt.edu (J. Xing), tw2@uab.edu (T. Wang), lixj04@gmail.com (X. Li), mxu1@cs.cmu.edu (M. Xu).

[1] These authors contributed equally to this work.

Different from most existing works, this paper reconsiders the fundamental design of the Mean-Teacher framework. Our work is motivated by the following two key limitations of existing MT-based approaches and their adverse effects on pseudo-label reliability. First, the strong correlation between the teacher and student networks induced by the EMA update process causes the teacher to inherit the student's instantaneous biases and prediction errors, leading to pseudo-labels that are highly self-reinforcing rather than corrective. As a result, erroneous or overconfident predictions from the student are repeatedly propagated and amplified, reducing the diversity and reliability of the generated pseudo-labels. Second, the static nature of EMA weights prevents the teacher from adaptively responding to rapid changes in the student's learning dynamics, causing a temporal mismatch between the teacher's guidance and the student's current representation quality. This lag effect further degrades pseudo-label quality, particularly during early training stages or periods of unstable optimization, where outdated teacher predictions can misguide the student and hinder robust learning.

To address these issues, we propose generalized improvements to the basic Mean-Teacher architecture that enhances the reliability of the teacher-student learning process, ensuring more robust and adaptive knowledge transfer. Specifically, we adopt a Dual-Student architecture to enhance diversity in the knowledge transferred to the teacher and a Student Selection module to improve its reliability. We also introduce a novel strategy named Loss-Aware Exponential Moving Average (LA-EMA), which utilizes the student model's loss behavior to adjust the weights for updating the teacher model. Different from existing MT-based networks, where knowledge is transferred from a single student to the teacher, limiting model diversity, our method introduces a dual-student mechanism that jointly transfers complementary knowledge to the teacher. This design enhances representational diversity and mitigates self-reinforcing bias in pseudo-label generation. Furthermore, unlike prior approaches that rely on a static EMA weighting scheme, we propose a dynamic EMA formulation that adaptively modulates the teacher update according to the evolving learning behavior of each student. This adaptive strategy enables the teacher to selectively integrate more reliable student signals over time, leading to more robust and informative supervision. Our contribution can be summarized as follows:

- Our work addresses the fundamental limitations of the standard Mean-Teacher architecture by introducing an improved framework with adaptive knowledge transfer between teacher and student networks.
- We propose a novel Student Selection module within a Dual-Student architecture to identify and utilize the most suitable student based on entropy scores, ensuring the teacher acquires optimal knowledge at each iteration.
- We introduce a novel strategy named Loss-Aware Exponential Moving Average (LA-EMA), which leverages the loss behavior of student models to compute adaptive weights for EMA updates.
- Extensive experiments on medical benchmarks, such as LA and ACDC, demonstrate that our method achieves state-of-the-art performance. Our method is also shown beneficial for general image segmentation tasks.

## 2. Related work

**Semi-Supervised Medical Image Segmentation.** Previous works in semi-supervised learning have focused on *Consistency Regularization* [10,17], *Pseudo-Labeling* [18], and *Co-training* [19,20]. For example, MC-Net [12] introduced a dual-decoder architecture with soft pseudo-labels and cross-supervision to produce consistent, low-entropy predictions and enhance feature generalization. Similarly, SS-Net [11] utilized adversarial noise and class prototypes for input perturbation, improving class separation and leveraging unlabeled data. DC-Net [4] applied a dynamic threshold to partition predictions into consistent and inconsistent components, using cross-supervision to learn from both. Furthermore, ABD [13] proposed an advanced Copy-Paste augmentation method, ranking regions to address mixed perturbation limitations and enhance consistency learning. Despite these advancements, confirmation bias remains a challenge [5], where incorrect pseudo-labels on unlabeled data degrade training performance.

**Teacher-Student Architecture.** In recent years, Mean-Teacher-based methods have excelled in Semi-Supervised Medical Image Segmentation (SSMIS) by effectively generating pseudo-labels and achieving state-of-the-art performance across benchmarks. The Mean-Teacher framework [1] uses a teacher model to produce pseudo-labels for guiding a student model, while an exponential moving average (EMA) module transfers learned knowledge from the student to the teacher. Building on this, PLGCL [21] applied contrastive learning between isolated class pixels in teacher-generated pseudo-labels and student predictions for more accurate multi-class segmentation. BCP [22] tackled distribution mismatches between labeled and unlabeled data with a Copy-Paste augmentation technique integrated into the Mean-Teacher framework. Dual-Teacher [23] proposed dual-teacher framework, where each teacher gained distinct knowledge from the student model to diversify teacher's guidance. Similarly, AD-MT [6] introduced a dual-teacher architecture with distinct augmentations to mitigate confirmation bias. Many other studies [24–27] have also leveraged the capability of multi-student or multi-teacher to biomedical applications. However, these methods rely on a fixed-weight EMA formula, which can degrade the teacher's performance during iterations when the student learns poorly. To address this, our proposed method proposes an adaptive EMA formula, allowing the teacher model to dynamically acquire knowledge from the student and enhance overall performance.

## 3. Methodology

### 3.1. Preliminary

As many semi-supervised medical segmentation methods, let $\mathcal{D}_L = \{(x_l, y_l)\}$ and $\mathcal{D}_U = \{x_u\}$ represent for labeled and unlabeled training data. Here, $x_l$ and $x_u$ are denoted as the labeled and unlabeled training samples, $y_l$ is defined as the one-hot groundtruth for labeled sample $x_l$. First, a V-Net [28] model is pretrained on the labeled set, which serves as the initial weights for both the student and teacher in the main training stage. Then, the Dual-Student Cross-Sample CutMix module (Section 3.2) is applied to enhance the diversity and generalization of student-teacher learning. To ensure better student selection for exchanging knowledge to the teacher, the Student Selection module (Section 3.3), combined with the Loss-Aware Exponential Moving Average module (Section 3.4), is employed. Finally, we provide a theoretical analysis from the perspective of the PAC-Bayes generalization bound, which motivates our empirical designs of the student selection and LA-EMA modules, as mentioned in A. The overall structure of our method can be visualized in Fig. 1 and written step-by-step in Algorithm 3.

### 3.2. Dual-student cross-sample CutMix

To enhance the diversity and generalization of the teacher model, a dual-student architecture is employed in our methodology. Let $\mathcal{S}_1$ and $\mathcal{S}_2$ represent the first and second students in our pipeline, while $\mathcal{T}$ denotes the teacher guiding both students. Inspired by [22], we leveraged *Cross-Sample CutMix* augmentation into $x_l$ and $x_u$ on *Dual-Student* framework, which can be formulated as:

$$x^{l2u} = \mathcal{M} \odot x_l + (1 - \mathcal{M}) \odot x_u, \quad (1)$$

$$x^{u2l} = (1 - \mathcal{M}) \odot x_l + \mathcal{M} \odot x_u \quad (2)$$

where $x_l \in \mathcal{D}_L$ and $x_u \in \mathcal{D}_U$, $\odot$ is the Hadamard product function and $\mathcal{M} \in (0,1)^{W\times H\times D}$ is called a zero-centered mask. To generate pseudo-labels for the unlabeled samples, $x_u$ is passed through the teacher $\mathcal{T}$. However, raw pseudo-labels generated by $\mathcal{T}$ may contain significant noise, which can degrade the performance of the guided students. To address this, a simple post-processing function is applied to select the largest connected component from the raw pseudo-labels, producing optimized pseudo-labels $\hat{y}_u$. The mixed labels, combining annotated labels $y_l$ and pseudo-labels $\hat{y}_u$, can then be expressed as:

$$y^{l2u} = \mathcal{M} \odot y_l + (1 - \mathcal{M}) \odot \hat{y}_u, \quad (3)$$

$$y^{u2l} = (1 - \mathcal{M}) \odot y_l + \mathcal{M} \odot \hat{y}_u \quad (4)$$

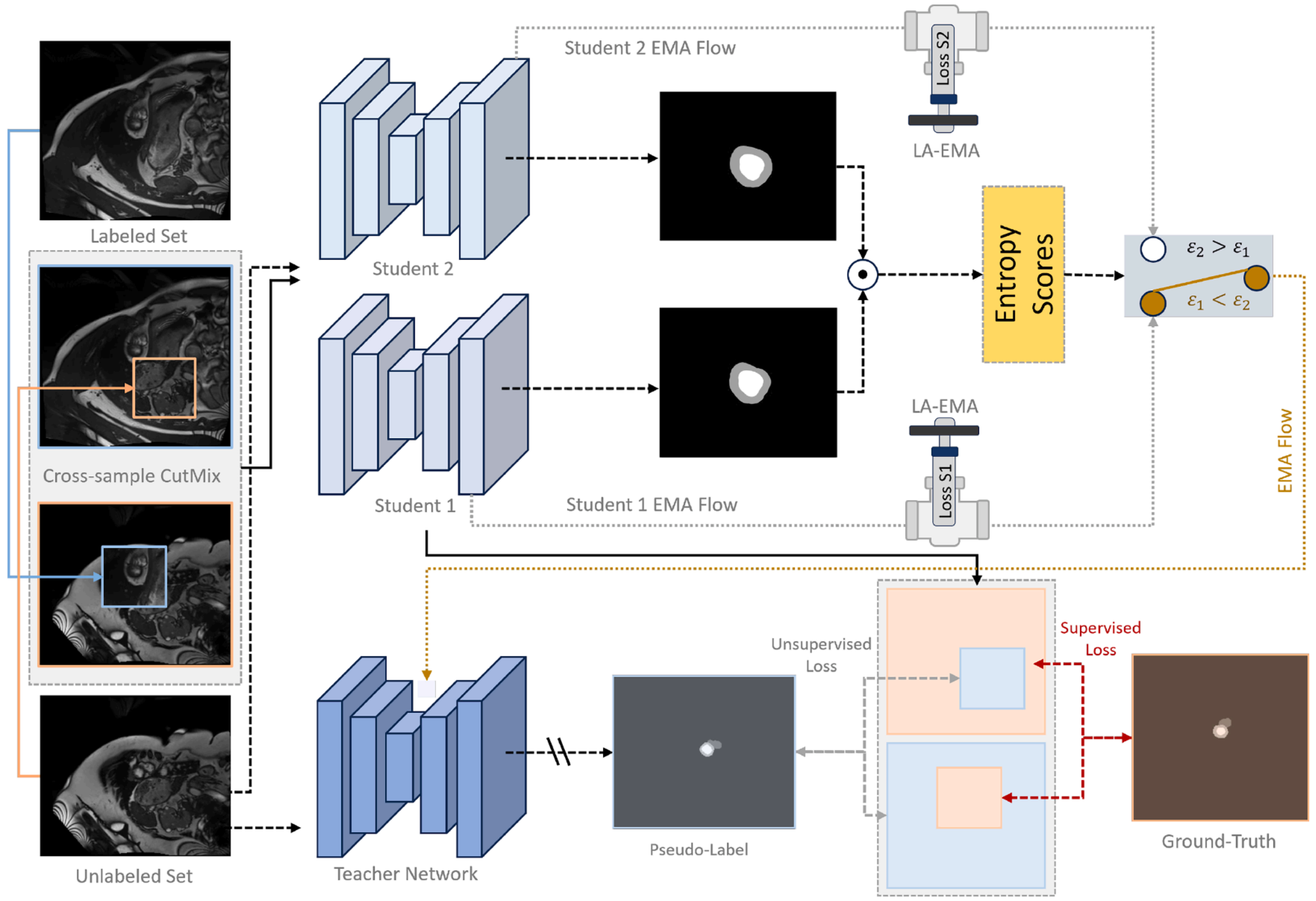

**Fig. 1.** The pipeline of our proposed framework consists of two parts. The training part involves the learning process of the dual-student model, where both the supervised and unsupervised losses are computed and optimized based on the predictions from the students and the pseudo-labels generated by the teacher through *Dual-Student Cross-Sample CutMix* module. The knowledge transfer part includes the *Student Selection* process, which identifies the best-performing student at each iteration and calculates its *Loss-Aware Exponential Moving Average (LA-EMA)* to update the teacher model. The $\odot$ represents for XNOR operation.

Both students, $\mathcal{S}_1$ and $\mathcal{S}_2$, are trained with two versions of the augmented samples, $x^{l2u}$ and $x^{u2l}$. These samples are passed through the students to produce predictions, $p_i^{l2u}$ and $p_i^{u2l}$, representing the predictions of the $i$th student for $i \in \{1, 2\}$. Finally, the loss function $\mathcal{L}^{\mathrm{CM}}$ is applied to both students and is formulated as:

$$\mathcal{L}_{CE,D}(p, y) = \mathcal{D}ice(p, y) + \mathcal{CE}(p, y), \quad (5)$$

$$\mathcal{L}_i^{CM} = \mathcal{L}_{CE,D}(p_i^{u2l}, y^{u2l}) + \mathcal{L}_{CE,D}(p_i^{l2u}, y^{l2u}) \quad (6)$$

where $\mathcal{D}ice(.)$ and $\mathcal{CE}(.)$ denote the Dice loss and Cross-Entropy loss functions, respectively. However, relying solely on $\mathcal{L}^{\mathrm{CM}}$ may lead to challenges. Specifically, the knowledge acquired by the students becomes highly dependent on this loss function. If $\mathcal{L}^{\mathrm{CM}}$ is insufficient to explore the full potential of the training data, the student's performance may stagnate at a local minimum.

To address this limitation and encourage students to extract more meaningful information from the training data, a support term is introduced. Let $\mathcal{M}^{l2u}$ and $\mathcal{M}^{u2l}$ represent the difference masks between the predictions of $\mathcal{S}_1$ and $\mathcal{S}_2$. Knowing that $\mathcal{M}^{l2u} = argmax(p_1^{l2u}) \oplus argmax(p_2^{l2u})$ and $\mathcal{M}^{u2l} = argmax(p_1^{u2l}) \oplus argmax(p_2^{u2l})$ where $\oplus$ denotes the XOR operation. Intuitively, the voxels or pixels with value 1 are those uncertain regions since the two students give inconsistent predictions. An additional loss term is introduced for both students:

$$\mathcal{L}_i^{MSE} = \mathrm{mean}(\mathcal{MSE}(p_i^{u2l}, y^{u2l}) \odot \mathcal{M}^{u2l}) + \mathrm{mean}(\mathcal{MSE}(p_i^{l2u}, y^{l2u}) \odot \mathcal{M}^{l2u}) \quad (7)$$

where $\mathcal{MSE}(.)$ denotes the Mean-Square Error (MSE) loss function. Using this additional term allows students to focus more on learning from uncertain regions, refining predictions to accurate voxels, and reducing model bias. The final loss function for optimizing each student is defined as:

$$\mathcal{L}_i = \alpha \cdot \mathcal{L}_i^{CM} + \beta \cdot \mathcal{L}_i^{MSE} \quad (8)$$

where $\alpha$ and $\beta$ are the respective weights for the two-loss components.

### 3.3. Student selection

During EMA process, one of the two students is selected to transfer the learned knowledge to the teacher using the proposed *Student Selection* algorithm. The Student Selection module addresses the varying confidence levels of dual-student models by ensuring that only the more reliable student updates the teacher. By aligning predictions through a consistency mask and selecting the student with lower entropy, this approach enhances the stability and quality of knowledge transfer. Initially, both students are evaluated on the unlabeled dataset $x_u$ to generate predictions $p_1^u$ and $p_2^u$, as follows:

$$p_1^u = \mathcal{S}_1(x_u), \qquad p_2^u = \mathcal{S}_2(x_u) \quad (9)$$

Next, the aligned mask $\mathcal{M}_u$ is computed as the intersection of these predictions:

$$\mathcal{M}_u = argmax(p_1^u) \odot argmax(p_2^u) \quad (10)$$

where $\odot$ denotes the XNOR operation, indicating regions where the predictions of the two students are consistent, for example, if a pixel is assigned the same class by both student models, its value is set to 1; otherwise, it is set to 0.

Finally, the entropy score is calculated for both students to evaluate their confidence in the predictions. The entropy scores are defined as:

$$\mathcal{E}_1 = \mathrm{mean}(\mathcal{CE}(p_1^u, argmax(p_1^u)) \odot \mathcal{M}_u), \quad (11)$$

$$\mathcal{E}_2 = \text{mean}(\mathcal{CE}(p_2^u, argmax(p_2^u)) \odot \mathcal{M}_u) \tag{12}$$

where the aligned mask $\mathcal{M}_u$ ensures the entropy is evaluated only on the mutual regions. The student with the smaller entropy score is selected to transfer their learned knowledge to the teacher. After selection process, the chosen student is taken into the LA-EMA module. The pseudo-code for Student Selection process is presented in Algorithm 1.

### 3.4. Loss-aware exponential moving average

To begin with, the standard Exponential Moving Average (EMA) formulation between the teacher and the student can be expressed as:

$$\theta_t^T = (1 - w) \cdot \theta_{t-1}^T + w \cdot \theta_t^S \tag{13}$$

where $w$ denotes the weight assigned to the student in the EMA process, and $\theta_t^T$, $\theta_t^S$ are the parameters of the teacher and the student at iteration $t$.

Directly using the chosen student to update the teacher via EMA can destabilize the teacher's parameters due to the student's potential noise and rapid changes. Prior works addressed this by assigning a fixed EMA weight $w$, emphasizing that late-iteration students, which had learn more knowledge, should influence the teacher more. However, poorly optimized late-iteration students can still degrade the teacher's performance. To mitigate this, our algorithm decomposes the standard EMA formula into two components. First, following prior approaches, the global EMA weight $w_t^{global}$ at iteration $t$ is defined as:

$$w_t^{global} = max(\frac{1}{1+t}, w_{max}) \tag{14}$$

This ensures that students in later iterations, having gained more knowledge, are assigned higher weights, while early-iteration students contribute less.

Second, $w_t^{decay}$ is additionally involved in modulating the contribution of the current student based on its loss behavior. Decay weight is computed as:

$$w_t^{decay} = \frac{1}{e^{\lambda \cdot loss_t}} \tag{15}$$

where $loss_t$ refers to the selected student's loss $\mathcal{L}_i$ at each iterations $t$, $\lambda$ is a constraint to balance the value range of loss. The term $w_t^{decay} \in (0, 1)$ ensures that students with smaller losses have smaller decay weights, allowing them to contribute more effectively. The final weight for the student at iteration $t$ is then computed by integrating both global and decay weight as:

$$w_t = max(\frac{1}{1+t}, w_{max}) \cdot \frac{1}{e^{\lambda \cdot loss_t}} \tag{16}$$

Finally, the standard EMA formula is updated using the computed weight to transfer knowledge from the selected student to the teacher:

$$\theta_t^T = (1 - w_t) \cdot \theta_{t-1}^T + w_t \cdot \theta_t^S \tag{17}$$

LA-EMA integrates both global and decay weights to ensure stable and effective learning for the teacher. Its pseudo-code is shown in the Algorithm 2.

**Algorithm 1** Student Selection.

**Input**: Student $\mathcal{S}_1$, Student $\mathcal{S}_2$, Unlabeled data $x_u$
**Output**: Chosen Student $\mathcal{S}$, Loss $\mathcal{L}$

1: Predict $p_1^u = \mathcal{S}_1(x_u)$.
2: Predict $p_2^u = \mathcal{S}_2(x_u)$.
3: Calculate $\mathcal{M}_u$ by Eq. (10).
4: Calculate $\mathcal{E}_1$ by Eq. (11).
5: Calculate $\mathcal{E}_2$ by Eq. (12).
6: **if** $\mathcal{E}_1 > \mathcal{E}_2$ **then**
7: $\quad \mathcal{S} = \mathcal{S}_2$, $\mathcal{L} = \mathcal{L}_2$
8: **else**
9: $\quad \mathcal{S} = \mathcal{S}_1$, $\mathcal{L} = \mathcal{L}_1$
10: **end if**
11: **return** $\mathcal{S}$, $\mathcal{L}$

**Algorithm 2** LA-EMA.

**Input**: The chosen student $\mathcal{S}$, loss $\mathcal{L}$, the iteration $t$
**Output**: The EMA weight $w$

1: Calculate $w^{global}$ by Eq. (14).
2: Calculate $w^{decay}$ by Eq. (15).
3: Calculate $w$ by Eq. (16).
4: **return** $w$

**Algorithm 3** Overview of the our proposed framework.

**Require:** Labeled data $\mathcal{D}_L$, unlabeled data $\mathcal{D}_U$
1: Pretrain a model $\theta$ on $\mathcal{D}_L$
2: Initialize teacher $\mathcal{T}$ and two students $\mathcal{S}_1, \mathcal{S}_2$ with the pretrained weight $\theta$
3: **for** each training iteration $t = 1, \ldots, T$ **do**
4: Sample labeled data $(x_l, y_l)$ and unlabeled data $x_u$
5: Mix labeled and unlabeled samples using **Cross-Sample CutMix** strategy.
6: Train both students on mixed samples and labels
7: Identify regions with inconsistent predictions between $\mathcal{S}_1$ and $\mathcal{S}_2$
8: Encourage students to refine predictions on uncertain regions
9: Select the student for teacher update via **Student Selection** mechanism
10: Update teacher parameters using **Loss-Aware EMA**
11: Evaluate on teacher $\mathcal{T}$ and two students $\mathcal{S}_1, \mathcal{S}_2$ to select the target model $\theta_{target}$
12: **end for**
13: **return** The target model $\theta_{target}$

## 4. Experiments

### 4.1. Experimental settings

**Datasets.** We use two public datasets as the benchmark for algorithm evaluation. **LA** [34] consists of 100 3D gadolinium-enhanced magnetic resonance image (GE-MRI) scans. Following previous works [6,22], the fixed data split for the training and validation set is 80 and 20 scans, respectively. **ACDC** [35] comprises 200 annotated short-axis cardiac cine-MR images from a cohort of 100 patients with four classes (i.e., background, right ventricle, left ventricle, and myocardium). The data split contains 70, 10, and 20 patients' scans for training, validation, and testing, respectively. To showcase the robustness of our proposed Loss-Aware EMA, we extend our evaluation beyond two medical image segmentation datasets to include two well-known general image datasets: CIFAR-10 and Pascal VOC. **CIFAR-10** [36] consists of 60,000 images in 10 classes (50,000 train, 10,000 validation), each of size $32 \times 32$. For our semi-supervised setting, 2% of the training data (1,000 images) is labeled and 98% (49,000) is unlabeled, as shown in Table 5. **Pascal VOC.** [37] contains diverse natural scenes with 20 object categories, posing challenges such as scale variation, occlusion, and complex backgrounds. We use 366 labeled and 1464 unlabeled images (20% labeled). Performance is evaluated with Intersection over Union (IoU) for both student (S) and teacher (T) networks.

**Metrics.** Four main metrics were selected for both datasets (LA and ACDC): Dice Similarity Coefficient (Dice), Jaccard Coefficient (Jaccard), 95% Hausdorff Distance (95HD), and Average Surface Distance (ASD). In addition, Accuracy (Acc) and Intersection over Union (IoU) metrics were chosen for general datasets.

**Implementation Details.** All experiments were conducted on an NVIDIA RTX A6000 GPU. The hyperparameter $\lambda$ and $w_{\max}$ were set to 0.3 and 0.01 respectively. Both $\alpha$ and $\beta$ were set to 0.5. The training process consisted of two steps: a pre-training step for weight initialization and a self-training step for implementing the main methodology. For training with LA dataset, the pre-training step consisted of 2000 iterations, followed by 15,000 iterations in the self-training step. The batch size was set to 8, with 4 samples allocated to the labeled set and 4 to the unlabeled set. Stochastic Gradient Descent (SGD) was used as the optimizer for both student models, with a learning rate of 0.01, momentum of 0.9, and weight decay of $1 \times 10^{-4}$. The input patch size was set to $112 \times 112 \times 80$, and V-Net 3D was used as the backbone. Data augmentations included random cropping and random flipping. Additionally, evaluation was performed every 200 iterations. For ACDC, the pre-training step consisted of 10,000 iterations, followed by 30,000 iterations in the self-training step. The Adam optimizer was employed for training, with a learning rate of 0.001. The batch size was set to 24, evenly split between 12 samples from the labeled set and 12 from the unlabeled set. U-Net 2D was used as the backbone, with an input patch size of $256 \times 256$. The evaluation process is set up the same as LA dataset training. For both settings, we set random seed at 1337 for implementation.

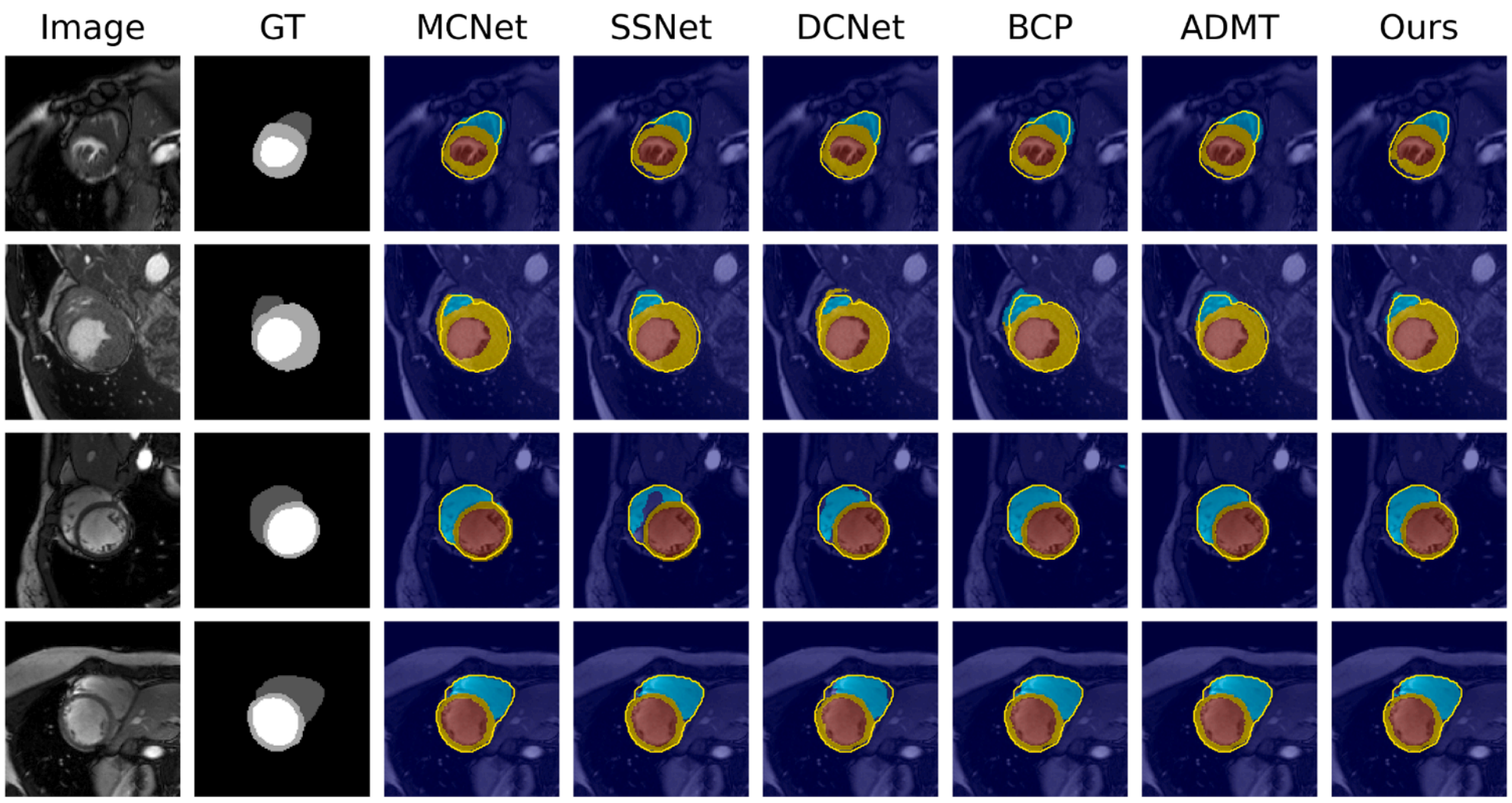

**Fig. 2.** Qualitative results of several samples from the ACDC test set.

**Table 1**
Comparison between our framework and various existing methods on the LA predefined test set. **Bold** and Underline represent the highest and second-highest number in our experiments.

| Method | Scans used | | Metrics | | | |
|---|---|---|---|---|---|---|
| | Labeled | Unlabeled | Dice ↑ | Jaccard ↑ | 95HD ↓ | ASD ↓ |
| V-Net [28] | 4 (5%) | 0 | 52.55 | 39.60 | 47.05 | 9.87 |
| | 8 (10%) | 0 | 82.74 | 71.72 | 13.35 | 3.26 |
| | 80 (100%) | 0 | 91.47 | 84.36 | 5.48 | 1.51 |
| URPC (MICCAI'21) [3] | 4 (5%) | 76 (95%) | 82.48 | 71.35 | 14.65 | 3.65 |
| MC-Net (MICCAI'21) [12] | | | 83.59 | 72.36 | 14.07 | 2.70 |
| SS-Net (MICCAI'22) [11] | | | 86.33 | 76.15 | 9.97 | 2.31 |
| PS-MT (CVPR'22) [2] | | | 88.49 | 79.13 | 8.12 | 2.78 |
| BCP (CVPR'23) [22] | | | 88.02 | 78.72 | 7.90 | 2.15 |
| AD-MT (ECCV'24) [6] | | | 89.63 | 81.28 | 6.56 | **1.85** |
| EIC (TMI'24) [29] | | | 87.35 | 77.66 | 8.96 | 2.02 |
| W2SPC (MedIA'25) [7] | | | 89.02 | 79.83 | 10.23 | 2.18 |
| PAMT (PR'25) [30] | | | 88.45 | 79.41 | 7.85 | 2.14 |
| Ours | | | **89.64** | **81.31** | **5.88** | 1.91 |
| URPC (MICCAI'21) [3] | 8 (10%) | 72 (90%) | 85.01 | 74.36 | 15.37 | 3.96 |
| SS-Net (MICCAI'22) [11] | | | 88.55 | 79.62 | 7.49 | 1.90 |
| MC-Net (MICCAI'21) [12] | | | 88.96 | 80.25 | 7.93 | 1.86 |
| PS-MT (CVPR'22) [2] | | | 89.72 | 81.48 | 6.94 | 1.92 |
| BCP (CVPR'23) [22] | | | 89.62 | 81.31 | 6.81 | 1.76 |
| AD-MT (ECCV'24) [6] | | | 89.83 | 81.62 | 6.99 | 1.70 |
| EIC (TMI'24) [29] | | | 89.25 | 80.68 | 6.96 | 1.86 |
| W2SPC (MedIA'25) [7] | | | 90.23 | 81.52 | 7.16 | 1.95 |
| PAMT (PR'25) [30] | | | 90.33 | 82.63 | 6.09 | 1.61 |
| Ours | | | **91.31** | **84.07** | **5.30** | **1.60** |

**Table 2**
Quantitative results on the ACDC test set comparing the proposed method with state-of-the-art methods, evaluated under two distinct settings: 5% and 10% labeled training samples.

| Method | Scans used | | Metrics | | | |
|---|---|---|---|---|---|---|
| | Labeled | Unlabeled | Dice ↑ | Jaccard ↑ | 95HD ↓ | ASD ↓ |
| U-Net [31] | 3 (5%) | 0 | 47.83 | 37.01 | 31.16 | 12.62 |
| | 7 (10%) | 0 | 79.41 | 68.11 | 9.35 | 2.70 |
| | 70 (100%) | 0 | 91.44 | 84.59 | 4.30 | 0.99 |
| MC-Net (MICCAI'21) [12] | 3 (5%) | 67 (95%) | 62.85 | 52.29 | 7.62 | 2.33 |
| SS-Net (MICCAI'22) [11] | | | 65.83 | 55.38 | 6.67 | 2.28 |
| DC-Net (MICCAI'23) [4] | | | 71.57 | 61.12 | 8.37 | 4.08 |
| BCP (CVPR'23) [22] | | | 87.59 | 78.67 | **1.90** | 0.67 |
| AD-MT (ECCV'24) [6] | | | 87.47 | 78.57 | 3.35 | 1.06 |
| EIC (TMI'24) [29] | | | 80.49 | 69.61 | 2.44 | **0.59** |
| DiffRect (MICCAI'24) [32] | | | 82.46 | 71.76 | 7.18 | 1.94 |
| W2SPC (MedIA) [7] | | | 82.39 | 71.16 | 5.67 | 2.01 |
| BoCLIS (TMI'25) [33] | | | 83.15 | 72.92 | 5.08 | 1.97 |
| Ours | | | **88.24** | **79.57** | 2.05 | 0.65 |
| MC-Net (MICCAI'21) [12] | 7 (10%) | 63 (90%) | 86.44 | 77.04 | 5.50 | 1.84 |
| SS-Net (MICCAI'22) [11] | | | 86.78 | 77.67 | 6.07 | 1.40 |
| DC-Net (MICCAI'23) [4] | | | 87.81 | 78.96 | 4.84 | 1.23 |
| BCP (CVPR'23) [22] | | | 88.84 | 80.62 | 3.98 | 1.17 |
| AD-MT (ECCV'24) [6] | | | 88.92 | 80.67 | 1.89 | 0.61 |
| EIC (TMI'24) [29] | | | 89.49 | 81.53 | 1.42 | 0.46 |
| DiffRect (MICCAI'24) [32] | | | 89.27 | 81.13 | 3.85 | 1.00 |
| W2SPC (MedIA'25) [7] | | | 88.92 | 79.65 | 4.32 | 1.19 |
| BoCLIS (TMI'25) [33] | | | 88.96 | 79.72 | 4.23 | 1.21 |
| Ours | | | **90.67** | **83.41** | **1.30** | **0.44** |

### 4.2. Comparison with state-of-the-art methods

**Quantitative Results.** In Tables 1 and 2, we compare our methodology with various approaches for semi-supervised medical segmentation. Particularly, our approach achieves superior performance in both the 5% and 10% labeled data settings. Notably, our proposed method demonstrates outstanding performance across both the LA and ACDC datasets, with exceptional results in the 10% labeled data setting. On both datasets, our approach consistently outperforms state-of-the-art

**Table 3**
Ablation studies for the plug-and-play **LA-EMA** approach on both the LA and ACDC datasets.

| LA (8/80 labeled data) | | |
|---|---|---|
| Method | Metrics | |
| | Dice ↑ | Jaccard ↑ |
| BCP (Reproduced) | 89.83 | 81.67 |
| + LA-EMA | **90.28** | **82.36** |
| Mean-Teacher | 88.68 | 79.81 |
| + LA-EMA | **89.18** | **80.60** |
| LA (16/80 labeled data) | | |
| Method | Metrics | |
| | Dice ↑ | Jaccard ↑ |
| UA-MT (Reproduced) | 88.06 | 78.92 |
| + LA-EMA | **88.52** | **79.68** |
| ACDC (7/70 labeled data) | | |
| Method | Metrics | |
| | Dice ↑ | Jaccard ↑ |
| BCP (Reproduced) | 89.25 | 81.16 |
| + LA-EMA | **89.84** | **82.12** |

**Table 4**
Ablation studies for our **Student Selection** module.

| ACDC (7/70 labeled data) | | |
|---|---|---|
| Dual-Student variants | Metrics | |
| | Dice ↑ | Jaccard ↑ |
| Single Student (1st) | 89.66 | 81.86 |
| Single Student (2nd) | 89.80 | 82.12 |
| Average of Student | 89.01 | 80.76 |
| Student Selection | **90.10** | **82.48** |

**Table 5**
Ablation Studies of **LA-EMA** on more general datasets.

| CIFAR-10 (1000/50000 labeled data) | | |
|---|---|---|
| Method | Metrics | |
| | Acc ↑ | |
| Mean-Teacher (Reproduced) | 82.36 | |
| + LA-EMA | **84.08** | |
| Pascal VOC (366/1464 labeled data) | | |
| Method | Metrics | |
| | IOU (S) ↑ | IOU (T) ↑ |
| UniMatchV2 (Reproduced) | 86.07 | 86.00 |
| + LA-EMA | **86.80** | **86.81** |

**Table 6**
Ablation studies for the components in the Dual-Student framework on the ACDC dataset (7/70 labeled).

| **ACDC (7/70 labeled data)** | | | | | |
|---|---|---|---|---|---|
| Base | CSC | SS | LA-EMA | Dice ↑ | Jaccard ↑ |
| ✓ | ✗ | ✗ | ✗ | 84.18 | 73.71 |
| ✓ | ✓ | ✗ | ✗ | 89.66 | 81.86 |
| ✓ | ✓ | ✓ | ✗ | 90.10 | 82.48 |
| ✓ | ✓ | ✓ | ✓ | **90.67** | **83.41** |

methods, showcasing its effectiveness in leveraging both labeled and unlabeled data. More specifically:

**LA Dataset.** In Table 1, we compare our methodology with various approaches for semi-supervised medical segmentation. Despite being a simple baseline, BCP (2023) demonstrates competitive performance on the LA dataset. Building upon [22], our approach achieves outstanding results under both the 5% and 10% labeled data training settings. Notably, with 10% labeled data, our method significantly outperforms all prior methods, achieving 91.31% in Dice similarity and 84.07% in the Jaccard score–surpassing the recently released AD-MT (reproduced) by 1.48% and 2.45%, respectively. Although our method trails AD-MT in the ASD metric for the 5% labeled data setting, it delivers the highest performance across all other metrics.

**ACDC Dataset.** Table 2 compares the performance of our method with other state-of-the-art (SOTA) approaches on the ACDC dataset using 5% and 10% labeled data. Our method surpasses SOTA methods across most metrics, except for the 95HD score in the 5% labeled data setting, where it ranks second. Under the 10% labeled data setting, our method achieves an 83.41% Jaccard score, outperforming AD-MT by 2.74%, demonstrating its ability to effectively leverage both labeled and unlabeled data. In the 5% labeled data setting, our method also outperforms SOTA methods, though with a smaller margin, indicating it may not fully utilize the available information when labeled data is very limited. This highlights the need for future research to develop strategies for better handling scenarios with sparse labeled data.

**Qualitative Results.** In Fig. 2, our method produces segmentation masks that closely match the ground truth compared to other methods. This visualization highlights the robustness and adaptability of our method, making it a powerful framework for semi-supervised medical segmentation. Fig. 3 shows the qualitative results on the LA dataset, comparing segmentation outputs from different methods. Notably, our method produces smoother and more accurate boundaries, closely aligning with the ground truth while reducing false positives and small disconnected regions. In contrast, BCP and AD-MT exhibit minor misalignments and occasional segmentation artifacts. The results emphasize the effectiveness of combining between switching dual-student and Loss-Aware EMA in the LA dataset in improving segmentation quality, particularly in addressing challenges such as anatomical variability, low contrast between tissues and noise in cardiac MRI images.

### *4.3. Ablation studies*

**Loss-aware Exponential Moving Average.** Table 3 compares three methods to evaluate the effectiveness of the proposed LA-EMA. Both naive BCP and UA-MT were tested with the standard EMA and LA-EMA modules, and LA-EMA consistently achieved higher performance. In particular, for BCP on the ACDC and LA datasets, LA-EMA outperformed the naive EMA. While LA-EMA consistently improves performance on medical imaging benchmarks (e.g., ACDC and LA), we further demonstrate that LA-EMA is a fully plug-and-play module that generalizes beyond medical imaging and can be integrated into diverse semi-supervised learning frameworks without architectural modification. In the experiments on CIFAR-10 with 2% labeled data (Table 5), replacing the standard EMA with the proposed LA-EMA led to a notable accuracy improvement, increasing from 82.36% to 84.08%, showcasing its effectiveness in leveraging unlabeled data. Similarly, on the Pascal VOC dataset with 20% labeled data, integrating LA-EMA into the UniMatch-V2 [38] method enhanced the Intersection over Union (IoU) scores for both the student (S) and teacher (T) models by approximately 0.8%, demonstrating the robustness of LA-EMA in handling complex segmentation tasks. These results demonstrate that LA-EMA functions as a task-agnostic, plug-and-play replacement for standard EMA across classification and segmentation tasks, highlighting its broad applicability beyond medical imaging.

**Student Selection.** Table 4 presents four experiments on the dual-student baseline using the ACDC dataset with 10% labeled data and the standard EMA formula. The "Single Student" approach involves the teacher learning from only one student during training. Results show minimal differences between using the first or second student individually, suggesting both contribute equally to teacher performance. The "Average Student" method, which transfers the mean of both students' weights to the teacher, performs poorly, while our Student Selection significantly boosts performance.

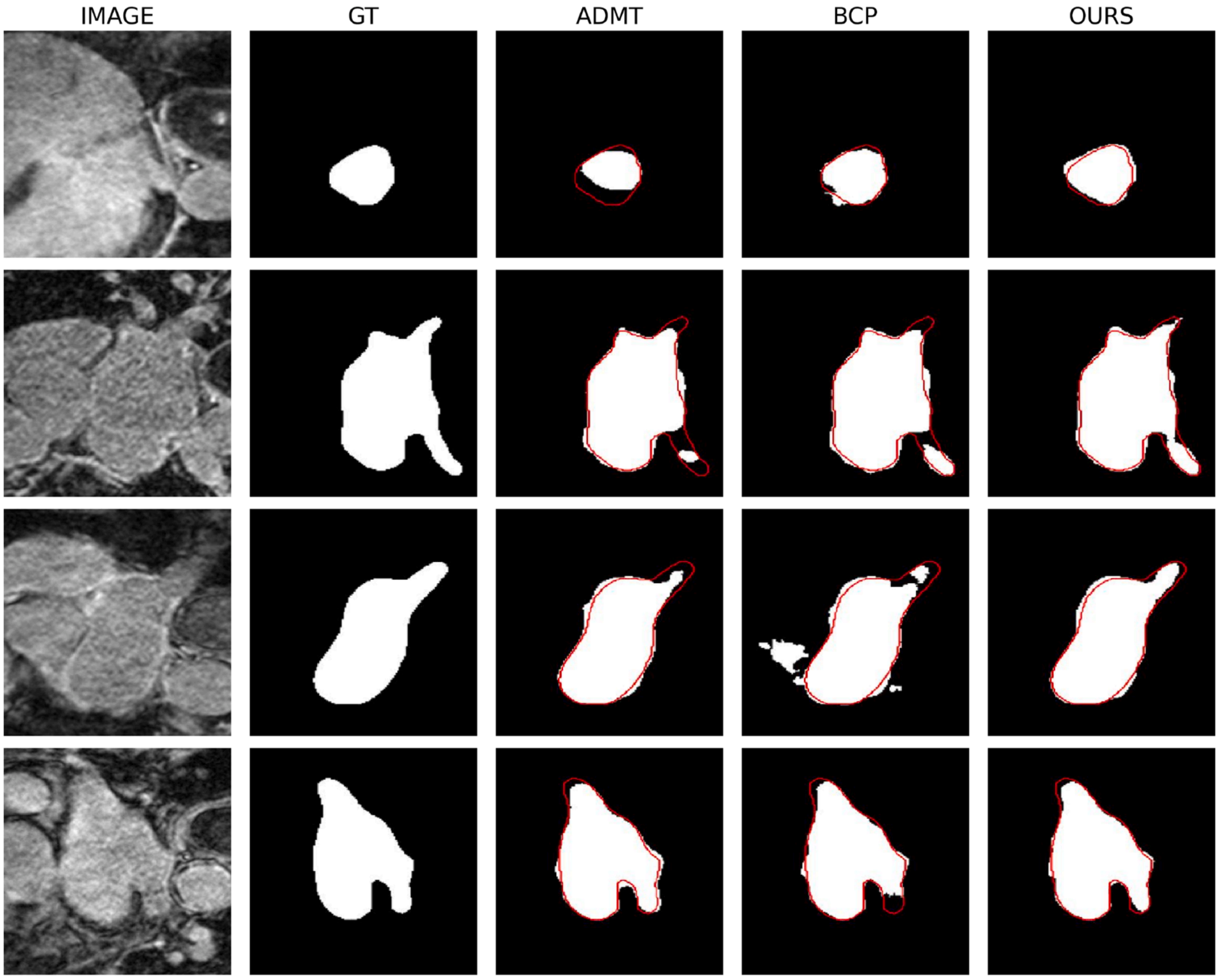

**Fig. 3.** Qualitative results on sample cases from the LA test set. From left to right, the columns display the Image, Ground Truth (GT), and predictions from various state-of-the-art methods, including BCP, AD-MT, and Ours.

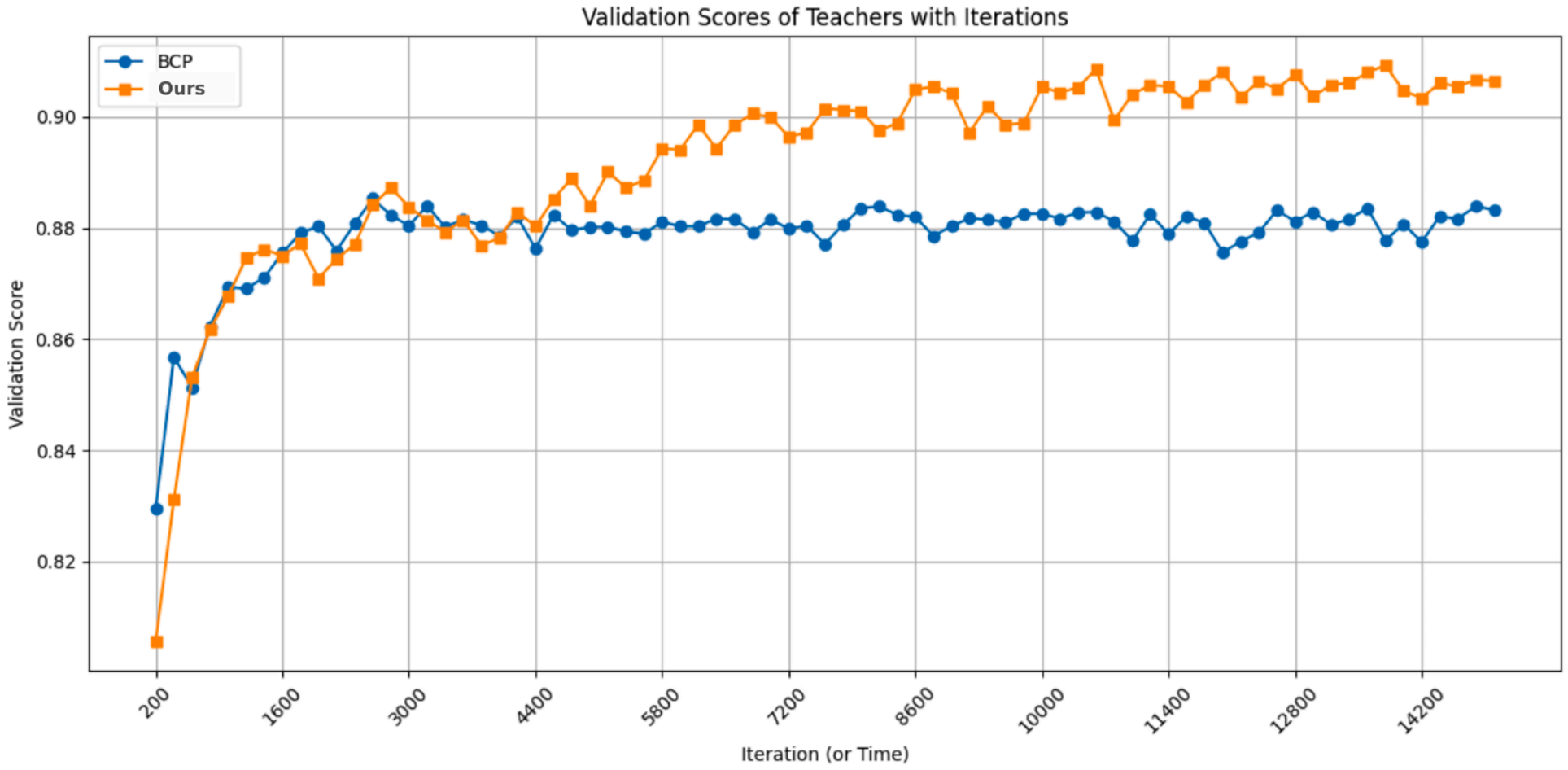

**Fig. 4.** The validation score (DICE) of teacher model of BCP and Ours over training.

**Component Analysis.** To evaluate each component of our framework, we conducted ablation experiments on the ACDC dataset with 10% of labeled data (Table 6). Starting from a base model, we observed lower performance. Adding the Cross-Sample CutMix (CSC) augmentation notably improved results, highlighting the benefit of consistency learning. Incorporating the Student Selection (SS) module further enhanced segmentation accuracy by selecting the most suitable student for the EMA process. Finally, adding LA-EMA yielded the best performance, confirming that transferring an appropriate amount of knowledge enhances teacher guidance. By jointly selecting the most suitable student and adaptively controlling the EMA update magnitude, LA-EMA prevents unreliable knowledge transfer and strengthens teacher supervision. These results demonstrate that the combination of dual-student selection and loss-aware EMA, rather than either component alone, con-stitutes the core innovation of our framework, with each module playing a complementary and mutually reinforcing role.

**Table 7**
Additional results on the ACDC test set comparing the proposed method with other methods under setting: 136/1049 labeled training slices.

| Method | Dice ↑ | Jaccard ↑ | Time | GPU |
|---|---|---|---|---|
| MC-Net (MICCAI'21) [12] | 88.44 ± 0.61 | 79.99 ± 0.85 | Appx 5h | Appx 5.7GB |
| SS-Net (MICCAI'22) [11] | 88.63 ± 0.31 | 80.22 ± 0.48 | Appx 4h | Appx 8.7GB |
| DC-Net (MICCAI'23) [4] | 88.93 ± 0.20 | 80.79 ± 0.34 | Appx 3h | Appx 6.5GB |
| BCP (CVPR'23) [22] | 88.57 ± 0.47 | 80.31 ± 0.81 | Appx 4h | Appx 3.5GB |
| AD-MT (ECCV'24) [6] | 88.96 ± 0.21 | 80.82 ± 0.32 | Appx 4h | Appx 5.9GB |
| Ours | **89.25** ± 0.26 | **81.31** ± 0.44 | Appx 7h | Appx 4.8GB |

**Comparison with BCP.** We further compare our method with BCP, since our approach integrates Cross-Sample CutMix. Fig. 4 shows that BCP's teacher stagnates in a local minimum mid-training, failing to exceed its performance ceiling. In contrast, our method steadily improves the teacher model throughout training. These findings underscore the effectiveness of combining Dual-Student and LA-EMA to enhance pseudo-label quality.

**Cross-validation Discussion.** We employed k-fold cross-validation and trained the model using three folds. As shown in Table 7, our method achieves state-of-the-art performance compared with existing frameworks, while exhibiting substantially lower variability. In particular, the standard deviations of the Dice and Jaccard metrics are reduced to 0.26% and 0.44%, respectively, indicating that our method delivers stable and consistent performance. We also compare the computational time and resource requirements of six methods, as summarized in Table 7. Although our method requires a longer training time, it delivers superior segmentation accuracy and localization performance, while exhibiting improved stability and generalizability on medical imaging data. This additional computational cost stems from the dual-student training strategy with knowledge transfer to a teacher network. In contrast, BCP employs a single-student design, which explains its lower GPU usage. Despite this, our method maintains a moderate GPU memory footprint compared with other competing approaches, while still achieving stronger overall performance.

**Discussion and Limitations.** While the proposed framework demonstrates consistent improvements across multiple benchmarks, several practical considerations are worth noting. First, the entropy-based student selection and Loss-Aware EMA mechanisms are designed to reduce the impact of unreliable pseudo-labels by suppressing high-uncertainty or high-loss updates, which provides a degree of robustness to noisy unlabeled data and moderate domain shifts. Nevertheless, as with most semi-supervised approaches, extreme domain discrepancies or highly corrupted unlabeled samples may still affect pseudo-label quality.

Second, the dual-student architecture introduces additional memory overhead compared to single-student methods. However, as shown in our computational analysis, this overhead remains moderate and represents a reasonable trade-off given the observed gains in stability and segmentation accuracy.

Finally, we adopt common post-processing strategies (e.g., largest connected component selection) following prior semi-supervised medical segmentation work. Our focus in this study is on improving the reliability of teacher–student knowledge transfer, and we leave a more exhaustive analysis of alternative post-processing strategies and large-scale multi-center data variability to future investigations.

## 5. Conclusion

In this paper, we propose a method incorporating two novel plug-and-play modules: the Loss-Aware Exponential Moving Average (LA-EMA) and the Student Selection modules. These modules are embedded into a Dual-Teacher architecture, demonstrating their effectiveness by significantly boosting performance across various benchmarks, such as SSMIS. Our pipeline enhances student learning while enabling the teacher to acquire more meaningful knowledge from the student models. This results in the generation of more accurate pseudo-labels, effectively mitigating confirmation bias issues and achieving state-of-the-art performance on various 3D datasets.

## CRediT authorship contribution statement

**Hoang-Thien Nguyen:** Writing – review & editing, Writing – original draft, Visualization, Validation, Software, Resources, Methodology, Investigation, Formal analysis, Conceptualization; **Thanh-Huy Nguyen:** Writing – review & editing, Writing – original draft, Visualization, Validation, Supervision, Resources, Project administration, Methodology, Investigation, Conceptualization; **Ba-Thinh Lam:** Writing – original draft, Validation, Methodology, Conceptualization; **Vi Vu:** Writing – original draft, Methodology, Conceptualization; **Bach X. Nguyen:** Software, Resources, Formal analysis; **Jianhua Xing:** Supervision, Project administration, Funding acquisition; **Tianyang Wang:** Supervision, Investigation, Formal analysis, Conceptualization; **Xingjian Li:** Writing – original draft, Supervision, Project administration, Investigation, Conceptualization; **Min Xu:** Writing – review & editing, Writing – original draft, Supervision, Resources, Project administration, Investigation, Funding acquisition.

## Data availability

Data will be made available on request.

## Declaration of competing interest

The authors declare that they have no known competing financial interests or personal relationships that could have appeared to influence the work reported in this paper.

## Acknowledgment

This work was supported in part by U.S. NIH grant R35GM158094, and NSF grant MCB-2205148. We thank AI VIETNAM for supporting us with GPUs to conduct experiments.

## Appendix A. Theoretical analysis

We provide a theoretical justification for our proposed dual-student semi-supervised learning framework with Loss-Aware EMA (LA-EMA). Our goal is to demonstrate that LA-EMA reduces the injected noise variance, the dual-student entropy-based selection leads to reliable update directions, and both together result in tighter generalization bounds. To do so, we introduce clear mathematical definitions and formalize the analysis in the form of three propositions.

Let $\theta_T^{(t)} \in \mathbb{R}^d$ denote the parameters of the teacher model at training iteration $t$, and let $\theta^* \in \mathbb{R}^d$ denote the optimal model that minimizes true risk. Let $\theta_{S_i}^{(t)} \in \mathbb{R}^d$ be the parameters of student $S_i \in \{1, 2\}$, and $\varepsilon_i^{(t)} \in \mathbb{R}^d$ be the stochastic noise associated with student $i$. We denote the student loss at iteration $t$ as $\mathcal{L}_{S_i}^{(t)} \in \mathbb{R}_+$, and the noise covariance as $\Sigma_i^{(t)} = \mathrm{Cov}(\varepsilon_i^{(t)})$. The softmax output of student $i$ on input $x$ is denoted $p_i(x) = (p_i^1(x), \dots, p_i^C(x)) \in \Delta^C$, and the entropy $\mathcal{H}(p_i(x)) = -\sum_{c=1}^{C} p_i^c(x) \log p_i^c(x)$.

### A.1. Proposition 1: Variance suppression through loss-aware EMA

We first show that LA-EMA suppresses noise variance injected into the teacher during training. We model the selected student $S^*$ at iteration $t$ as a noisy estimator of $\theta^*$, such that $\theta_{S^*}^{(t)} = \theta^* + \varepsilon^{(t)}$, with $\mathbb{E}[\varepsilon^{(t)}] = 0$ and $\mathrm{Cov}[\varepsilon^{(t)}] = \Sigma^{(t)}$. The teacher update rule becomes:

$$\theta_T^{(t)} = (1 - w_t)\theta_T^{(t-1)} + w_t \theta_{S^*}^{(t)}. \tag{A.1}$$

To analyze the noise effect in isolation, suppose the teacher is currently optimal, i.e., $\theta_T^{(t-1)} = \theta^*$. Then:

$$\theta_T^{(t)} = (1 - w_t)\theta^* + w_t(\theta^* + \varepsilon^{(t)}) = \theta^* + w_t \varepsilon^{(t)}. \tag{A.2}$$

Hence, the deviation becomes:

$$\theta_T^{(t)} - \theta^* = w_t \varepsilon^{(t)}, \tag{A.3}$$

and the expected squared deviation is:

$$\begin{aligned} \mathbb{E}[\|\theta_T^{(t)} - \theta^*\|^2] &= \mathbb{E}[\|w_t \varepsilon^{(t)}\|^2] \\ &= w_t^2 \cdot \mathbb{E}[\|\varepsilon^{(t)}\|^2] \\ &= w_t^2 \cdot \mathrm{Tr}(\Sigma^{(t)}). \end{aligned} \tag{A.4}$$

Then, we compare $\mathbb{E}_{\mathrm{LA}}[\|\theta_T^{(t)} - \theta^*\|^2]$ with $\mathbb{E}_{\mathrm{EMA}}[\|\theta_T^{(t)} - \theta^*\|^2]$ to highlight the variance suppression of our LA-EMA, where the update weight of the standard EMA is decayed by training iterations $t$, $w_{EMA} = w_t^{global} = max(\frac{1}{1+t}, w_{\max})$. In LA-EMA, we define $w_t = w_t^{global} \frac{1}{\exp(\lambda \mathcal{L}_{S^*}^{(t)})} = w_{EMA} \frac{1}{\exp(\lambda \mathcal{L}_{S^*}^{(t)})}$. Suppose the student loss $\mathcal{L}_{S^*}^{(t)} > 0$, then it follows that $w_t < w_{EMA}$ for all iterations. Thus, under the same noise covariance $\Sigma^{(t)}$, the expected variance of LA-EMA is:

$$\mathbb{E}_{\mathrm{LA}}[\|\theta_T^{(t)} - \theta^*\|^2] = w_t^2 \cdot \mathrm{Tr}(\Sigma^{(t)}), \tag{A.5}$$

while the expected variance under standard EMA is:

$$\mathbb{E}_{\mathrm{EMA}}[\|\theta_T^{(t)} - \theta^*\|^2] = w_{EMA}^2 \cdot \mathrm{Tr}(\Sigma^{(t)}). \tag{A.6}$$

Since $w_t < w_{EMA}$, it follows that $w_t^2 < w_{EMA}^2$, and thus:

$$\mathbb{E}_{\mathrm{LA}}[\|\theta_T^{(t)} - \theta^*\|^2] < \mathbb{E}_{\mathrm{EMA}}[\|\theta_T^{(t)} - \theta^*\|^2]. \tag{A.7}$$

We conclude that LA-EMA injects less noise into the teacher than standard EMA, providing a variance-suppressing advantage that stabilizes the learning trajectory and reduces deviation from optimality.

### A.2. Proposition 2: Reliable descent via entropy-guided student selection

We now show that the entropy-based student selection mechanism encourages updates aligned with reliable and confident predictions. Let the mutual agreement region be:

$$M_u := \{x \in \mathcal{X} \mid \arg\max p_1(x) = \arg\max p_2(x)\}. \tag{A.8}$$

For each student $S_i$, define the total entropy over $M_u$ as:

$$E_i^{(t)} := \sum_{x \in M_u} \mathcal{H}(p_i(x)) = -\sum_{x \in M_u} \sum_{c=1}^{C} p_i^c(x) \log p_i^c(x). \tag{A.9}$$

The selected student is then given by:

$$S^* = \arg\min_{i \in \{1,2\}} E_i^{(t)}. \tag{A.10}$$

Since lower entropy indicates higher prediction confidence, this selection favors the student that is more certain in consensus regions. This reduces the probability of propagating incorrect pseudo-labels and encourages updates that descend the empirical loss landscape more reliably.

### A.3. Proposition 3: Generalization improvement under PAC-bayes bound

We conclude by connecting LA-EMA and entropy-based selection to generalization guarantees. For any $\delta \in (0, 1]$, with high probability $1 - \delta$, the PAC-Bayes bound on the generalization error $R(f_T)$ of the teacher, with posterior $\rho$ centered at $\theta_T^{(t)}$ and prior $\pi$, is:

$$R(f_T) \le \mathcal{L}(f_T) + \sqrt{\frac{KL(\rho\|\pi) + \log(1/\delta)}{2n}}. \tag{A.11}$$

The empirical loss $\mathcal{L}(f_T)$ is defined as:

$$\mathcal{L}(f_T) = \frac{1}{n} \sum_{i=1}^{n} \ell(f(x_i; \theta_T^{(t)}), y_i), \tag{A.12}$$

where $\ell$ is the supervised loss on the labeled dataset. From Proposition 1, LA-EMA reduces the variance of updates due to high-loss students, keeping $\theta_T^{(t)}$ closer to its prior trajectory and reducing $KL(\rho\|\pi)$. From Proposition 2, selecting the more confident student mitigates pseudo-label noise, thereby keeping $\mathcal{L}(f_T)$ low. Thus, both components act to tighten the PAC-Bayes bound:

$$R(f_T) \lessapprox \mathcal{L}(f_T)^{\mathrm{low}} + \sqrt{\frac{KL^{\mathrm{small}}}{2n}}. \tag{A.13}$$

### A.4. Summary

The propositions presented demonstrate that our method is theoretically grounded. Proposition 1 shows that LA-EMA scales down noisy student updates and minimizes the variance injected into the teacher. Proposition 2 shows that the entropy-based student selection filters uncertain pseudo-labels and aligns updates with high-confidence predictions. Proposition 3 links these mechanisms to improved PAC-Bayes generalization bounds by simultaneously lowering empirical loss and controlling posterior complexity. These results establish that our dual-student framework with LA-EMA promotes stable, generalizable learning under semi-supervised settings.